\documentclass{article} %
\usepackage{colm2024_conference}

\usepackage{booktabs}
\usepackage{graphicx}
\usepackage{enumitem}
\usepackage{wrapfig}
\usepackage{algorithm}
\usepackage{algpseudocode}
\usepackage{multicol}

\usepackage{microtype}
\usepackage{amsmath}
\usepackage{colortbl}
\usepackage[utf8]{inputenc}
\definecolor{lightgray}{rgb}{0.9,0.9,0.9}
\usepackage{caption}
\usepackage{subcaption}
\usepackage{xcolor}
\usepackage{setspace}
\usepackage{url}
\usepackage{multirow}
\usepackage{colortbl}
\usepackage{tabularx}
\usepackage{blindtext}
\usepackage{pgfplots}
\pgfplotsset{compat=1.18} 
\usepackage{tikz}
\usetikzlibrary{er,positioning,bayesnet}
\usepackage{makecell}
\usepackage{tipa}
\usepackage{siunitx}
\usepackage{nicefrac}
\usepackage{tocloft}
\usepackage{listings}
\usepackage[raster,skins]{tcolorbox} %
\usepackage{xltabular}
\usepackage{adjustbox}
\usepackage{xurl}
\usepackage{setspace}

\usepackage{lipsum}
\usepackage{arydshln} 

\usepackage{amsmath,amsfonts,bm}

\def\eqref#1{equation~\ref{#1}}

\def\1{\bm{1}}

\DeclareMathAlphabet{\mathsfit}{\encodingdefault}{\sfdefault}{m}{sl}
\SetMathAlphabet{\mathsfit}{bold}{\encodingdefault}{\sfdefault}{bx}{n}

\title{MarsRL: Advancing Multi-Agent Reasoning System \\via Reinforcement Learning with Agentic Pipeline Parallelism}

\author{
\text{Shulin Liu}$^*$, \text{Dong Du}$^*$, \text{Tao Yang}$^*$, \text{Yang Li}, \text{Boyu Qiu} \\
Tencent Hunyuan Team \\
\texttt{\{forestliu,dongdu,rigorosyang,youngyli,boyuqiu\}@tencent.com} \\
}

\begin{document}

\maketitle

\begin{center}
    \textsuperscript{*}\textit{Contribute equally to this work.}
\end{center}

\begin{abstract}
Recent progress in large language models (LLMs) has been propelled by reinforcement learning with verifiable rewards (RLVR) and test-time scaling. However, the limited output length of LLMs constrains the depth of reasoning attainable in a single inference process. Multi-agent reasoning systems offer a promising alternative by employing multiple agents including Solver, Verifier, and Corrector, to iteratively refine solutions. While effective in top closed-source models, they struggle to generalize to open-source models due to insufficient critic and correction capabilities. To address this, we propose MarsRL, a novel reinforcement learning framework with agentic pipeline parallelism, designed to jointly optimize all agents in the system. MarsRL introduces agent-specific reward mechanisms to mitigate reward noise and employs pipeline-inspired training to enhance efficiency in handling long trajectories. Experimental results show that MarsRL improves AIME2025 accuracy from 86.5\% to 93.3\% and BeyondAIME from 64.9\% to 73.8\% when applied to Qwen3-30B-A3B-Thinking-2507. These findings highlight the potential of MarsRL to advance multi-agent reasoning systems and broaden their applicability across diverse reasoning tasks. The trained model and the multi-agent reasoning system are publicly available\footnote{\url{https://github.com/liushulinle/MarsRL}}.
\end{abstract}

\begin{figure}[ht]
  \centering
  \includegraphics[width=0.8\linewidth]{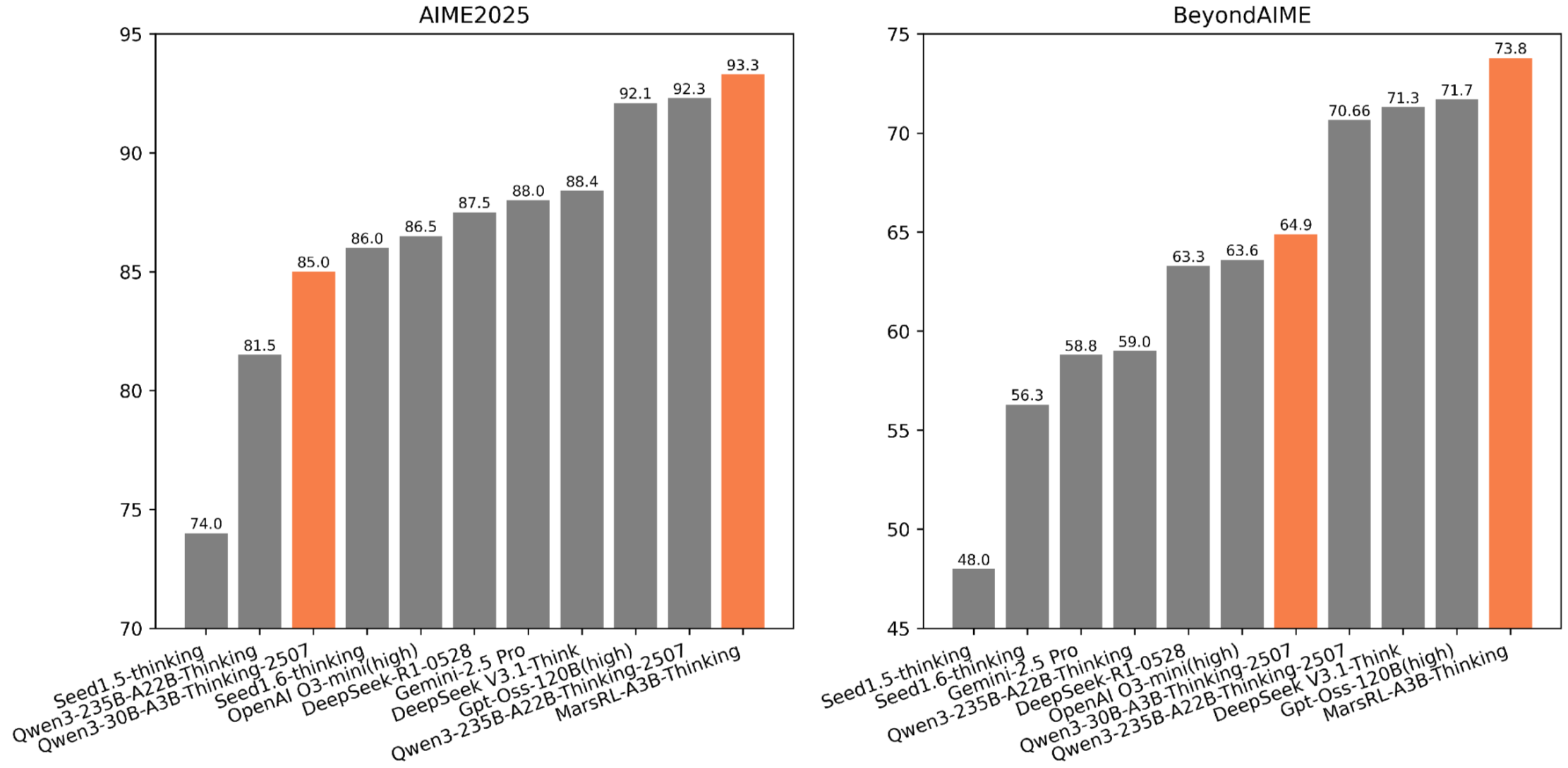}
  \label{fig:my_label}
\end{figure}

\section{Introduction}
Recent advancements in large language models (LLMs) have markedly improved their reasoning capabilities across complex domains such as mathematics and programming. This progress has been driven by state-of-the-art models, including OpenAI o1~\citep{jaech2024openai}, DeepSeek R1~\citep{guo2025deepseek}, and others that leverage test-time scaling strategies. A pivotal factor underlying their success is the integration of reinforcement learning with verifiable rewards (RLVR). In our prior work~\citep{du2025ulorl}, we demonstrated that progressively extending output length through RLVR results in sustained enhancements in the model's reasoning performance. However, the computational complexity of the Transformer architecture~\citep{vaswani2017attention} scales quadratically with sequence length, leading to a substantial increase in resource requirements and training time as output length grows. This constraint imposes a practical limitation on the maximum output length achievable by the model, thereby restricting the depth of reasoning attainable in a single inference process.

The Multi-Agent reasoning system adopts a novel approach to test-time scaling by leveraging multiple agents to perform iterative reasoning. This method enhances the depth of the model's reasoning process and addresses the limitations associated with the reasoning length of a single model~\citep{estornell2024acc, chai2025scimaster, estornell2025train, huang2025gemini}. Notably, \cite{huang2025gemini} successfully solved 5 out of 6 IMO-2025 problems by implementing a self-verification pipeline based on Gemini 2.5 Pro, achieving performance at a gold medal level. Their approach utilized iterative verification and refinement of the model's responses, employing multiple rounds to deepen the reasoning process. In contrast, Gemini 2.5 Pro, when restricted to single-round problem-solving, failed to correctly solve any of the problems. This stark comparison underscores the effectiveness of the Multi-agent reasoning system.

The pipeline described in \cite{huang2025gemini} aligns with a common human strategy for complex reasoning: generate an initial solution, check it for errors, and iteratively repair identified issues. Although their system was developed for the International Mathematical Olympiad (IMO), its structure is, in principle, applicable to general reasoning tasks. To promote broader applicability, we recast this process within a multi-agent framework (Figure \ref{fig:v-c agents}) comprising three roles: Solver, Verifier, and Corrector. The Solver produces the initial solution; the Verifier detects potential errors; and the Corrector amends the solution accordingly. Because iterative improvement is primarily driven by the Verifier and Corrector, we refer to this framework as the V–C reasoning system.

Although \cite{huang2025gemini} demonstrated that the Verifier–Corrector (V–C) reasoning paradigm is highly effective with Gemini 2.5 Pro, directly porting this system to recent open-source models, such as Qwen3-A22B-Thinking-2507 and DeepSeek V3.1-Think, does not yield improvements on AIME 2025 and BeyondAIME \citep{yu2025dapo}. We attribute this gap to the limited critiquing and correction capabilities of these open-source models. To make the paradigm effective in this setting, we employ agentic reinforcement learning to jointly optimize the Solver, Verifier, and Corrector, thereby improving cross-component coordination and overall performance.

Existing approaches to agentic reinforcement learning predominantly focus on Tool-Integrated Reasoning (TIR) models, which integrate external tools into a single LLM’s reasoning loop~\citep{chai2025scimaster, xue2025simpletir, dong2025agentic, shang2025rstar2}. In TIR, a single agent interacts with various tools. During reasoning, the agent invokes these tools to perform computations or retrieve external information, subsequently continuing its reasoning based on the feedback received. This paradigm fundamentally represents single-agent reasoning. Although MLPO~\citep{estornell2025train} involves multiple agents, its reinforcement training optimizes only a single agent while freezing the others. In contrast, our work addresses a multi-agent scenario where all agents are simultaneously optimized through reinforcement learning. This introduces two unique challenges:
\begin{itemize}
    \item \textbf{Reward Noise} In RL, the final reward is typically assigned to the entire trajectory. However, in multi-agent systems, this approach can introduce significant noise. For instance, consider a trajectory comprising three components: [$s$, $b$, $c$], where $s$ is the solution generated by the Solver, $b$ is the bug report from the Verifier, and $c$ is the refined result from the Corrector. Suppose $s$ is correct but the Verifier erroneously flags it as incorrect, and $c$ is ultimately correct, the reward for the entire trajectory is determined by $c$. Consequently, $b$ is assigned a positive reward despite the Verifier's incorrect judgment. This misalignment highlights the challenge of accurately attributing rewards in multi-agent systems.
    \item \textbf{Training Efficiency} Assuming the maximum output length of the LLM is $L$ and a single sample passes through up to $n$ agents, the trajectory length can reach $n \times L$. In our setting, with $n = 5$ and $L = 64$k, the trajectory length is up to 320k. Rolling out such long trajectories is computationally inefficient, particularly when dealing with the long-tail distribution of trajectory lengths.
\end{itemize}

To address these challenges, we propose a novel Agentic RL framework to advance \textbf{M}ulti-\textbf{A}gent \textbf{R}easoning \textbf{S}ystems via \textbf{R}einforcement \textbf{L}earning with Agentic Pipeline Parallelism (MarsRL). To mitigate the issue of reward noise, MarsRL assigns individualized rewards to each agent, thereby decoupling credit assignment across roles. Specifically, the Solver and the Corrector receive rewards computed directly from the agreement between their produced solutions and the reference answer. The Verifier’s reward, by contrast, is determined by the correctness of its judgment with respect to the actual correctness of a given solution: for example, if a solution matches the reference answer but the Verifier incorrectly labels it as erroneous, the Verifier is penalized. Moreover, to enhance training efficiency, we draw inspiration from pipeline parallelism ~\citep{huang2019gpipe}. Specifically, we decompose the training process into a pipeline at the agent level. Once an agent completes its decoding step, its output is immediately added to the training queue, eliminating the need to wait for the completion of the entire trajectory. This approach significantly reduces the time delay between trajectory generation and training.

We validated the effectiveness of the proposed method through a series of experiments. Specifically, we applied MarsRL to the Qwen3-30B-A3B-Thinking-2507 model. After training, the model’s performance on AIME2025 increased from 86.5\% to 93.3\%, and on BeyondAIME from 64.9\% to 73.8\%. Notably, it surpassed the larger Qwen3-235B-A22B-Thinking-2507 model (AIME2025: 92.3\%, BeyondAIME: 70.6\%). These results highlights the substantial gains achieved by our approach. Moreover, our extensive experiments demonstrate that the Verifier and Corrector trained via MarsRL generalize effectively across different Solvers.

\section{Preliminary}
\subsection{The Verifier-Corrector Reasoning System}
\begin{figure}[ht]
  \centering
  \includegraphics[width=0.9\textwidth]{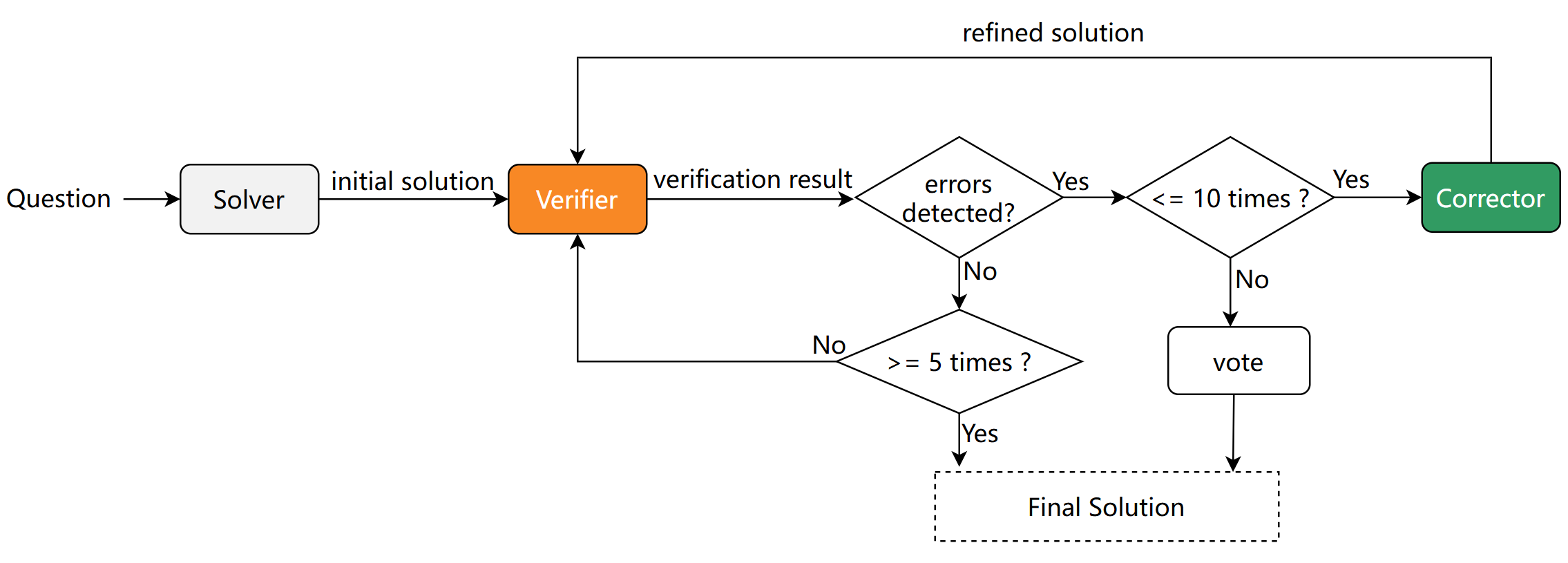}
  \caption{Overview of the Verifier-Corrector Reasoning System.}
  \label{fig:v-c agents}
\end{figure}

Recently, \cite{huang2025gemini} reported solving five of the six IMO 2025 problems using a self-verification pipeline built on Gemini 2.5 Pro, achieving gold-medal-level performance. To enhance generality and modularity, we reframe the pipeline from a multi-agent perspective (Figure \ref{fig:v-c agents}), comprising three agents: Solver, Verifier, and Corrector. The Solver produces the initial solution, while the Verifier diagnoses defects and the Corrector revises the solution accordingly. Relative to the original pipeline, the principal modifications are as follows:
\begin{itemize}
    \item For the Solver, the system prompt, which is intended to prioritize rigor over the pursuit of a definitive answer, was removed.
    \item \cite{huang2025gemini} introduced a refinement step to extend the maximum context length from 32k to 64k tokens. As recent models (e.g., Qwen3-A3B-Thinking-2507 and DeepseekV3.1) support context windows exceeding 64k tokens, we omit this step in our implementation.
    \item To ensure that every problem yields a final solution, if no candidate passes the acceptance check, the solution most frequently identified as correct by the Verifier is selected as the final solution.
\end{itemize}

\subsection{GRPO}
GRPO~\citep{shao2024deepseekmath} introduces a group-relative advantage estimation approach as an alternative to PPO~\citep{schulman2017proximal}, effectively removing the reliance on value functions. For any question-answer pair $(q,a)$, the behavioral policy $\pi_{\theta_{\text{old}}}$ generates a group of $G$ distinct responses $\{o_i\}_{i=1}^G$. The advantage for the $i$-th response $\hat{\mathcal{A}}_{i,t}$ is derived through group-level normalization:
\begin{equation}
\hat{\mathcal{A}}_{i,t} = \frac{r_i - \text{mean}(\{\mathcal{R}_i\}_{i=1}^G)}{\text{std}(\{\mathcal{R}_i\}_{i=1}^G)}
\end{equation}
GRPO also adopts a clipped surrogate function,  together with explicit KL regularization against a reference policy:
\begin{equation}
\begin{aligned}
\mathcal{J}_{\text{GRPO}}(\theta) = \mathbb{E}_{\substack{(q,a)\sim\mathcal{D}, \{o_i\}_{i=1}^G\sim\pi_{\theta_{\text{old}}}(\cdot|q)}}\\
\Biggl[ \frac{1}{G}\sum_{i=1}^G \frac{1}{|o_i|} \sum_{t=1}^{|o_i|} \biggl( \min\Bigl( & r_{i,t}(\theta) \hat{\mathcal{A}}_{i,t},  
 \text{clip}\bigl(r_{i,t}(\theta), 1-\varepsilon, 1+\varepsilon\bigr) \hat{\mathcal{A}}_{i,t} \Bigr) 
 - \beta D_{\text{KL}}\left(\pi_\theta \parallel \pi_{\text{ref}}\right) \biggr) \Biggr]
\end{aligned}
\end{equation}
where the importance ratio $r_{i,t}(\theta)$ measures policy update magnitude:
\begin{equation}
r_{i,t}(\theta) = \frac{\pi_{\theta}(o_{i,t} \mid q, o_{i,<t})}{\pi_{\theta_{\text{old}}}(o_{i,t} \mid q, o_{i,<t})}
\end{equation}

\subsection{UloRL}
In our recent study, we introduced a series of effective modifications to the GRPO framework, aimed at enhancing the efficiency of training for tasks involving ultra-long outputs~\citep{du2025ulorl}. These modifications are also incorporated into this work.

\textbf{Segment Rollouts} In tasks that require generating ultra-long sequences, the long-tail latency in rollouts becomes a dominant bottleneck. We mitigate this issue by staging the decoding process. At each stage, only a fixed-length segment is decoded. Trajectories that finish within a stage are immediately added to the experience replay buffer for training, whereas unfinished trajectories are carried forward to the next stage; their previously generated prefixes are concatenated with the new segment and decoding resumes.

\textbf{Dynamic Mask Well-mastered Positive Tokens} 
Entropy collapse is a significant challenge in reinforcement learning (RL) training \citep{cheng2025reasoning, he2025skywork, yu2025dapo, zhu2025surprising}. In UloRL, we identified the over-training of well-mastered positive tokens (MPTs) as the primary factor contributing to entropy reduction. To address this issue, we proposed a novel approach termed Dynamic Masking of MPTs. This method involves selectively masking MPTs during training, but only when the current entropy falls below a predefined target entropy threshold. Please refer to the full paper for algorithmic and implementation details\footnote{\url{https://arxiv.org/pdf/2507.19766}}.

\section{MarsRL}
\cite{huang2025gemini} demonstrated that the V-C reasoning system exhibits high effectiveness when applied to the Gemini 2.5 Pro. However, our findings indicate that this system fails to generalize to open-source models, such as Qwen3-A22B-Thinking-2507 and DeepSeek V3.1-Think. In this section, we introduce the proposed MarsRL framework, an agentic reinforcement learning framework that jointly trains the Solver, Verifier, and Corrector.

\subsection{Modeling the V-C Reasoning System in Agentic RL}
\begin{figure}[ht]
  \centering
  \includegraphics[width=0.9\linewidth]{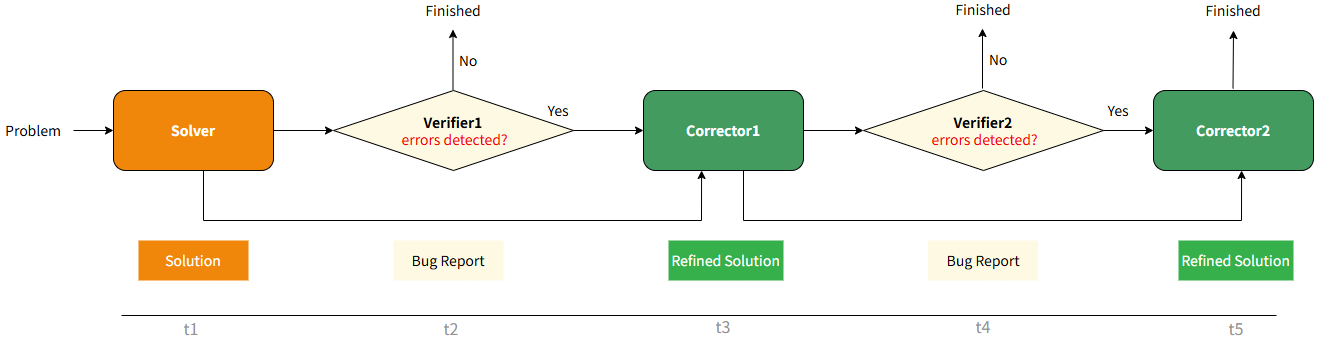}
  \caption{Modeling the V-C Reasoning System in Agentic RL.}
  \label{fig:rl_pipline}
\end{figure}
As depicted in Figure \ref{fig:rl_pipline}, the V-C reasoning process is sequentially unfolded across time steps $t_1$ to $t_5$:
\begin{itemize}
    \item $t_1$: The Solver generates an initial solution, denoted as \( s \), for the given problem.
    \item $t_2$: Verifier1 evaluates \( s \) and produces a bug report, denoted as \( br_1 \). If no errors are detected, the process terminates; otherwise, the process proceeds to the next step.
    \item $t_3$: Corrector1 refines \( s \) based on \( br_1 \), resulting in a refined solution, denoted as \( rs_1 \).
    \item $t_4$: Verifier2 evaluates \( rs_1 \) and generates another bug report, denoted as \( br_2 \). If no errors are detected, the process terminates; otherwise, the process advances to the next step.
    \item $t_5$: Corrector2 refines \( rs_1 \) based on \( br_2 \), producing a further refined solution, denoted as \( rs_2 \).
\end{itemize}
Verifier1 and Verifier2 operate under the same system prompt and template but process distinct types of data. Verifier1 assesses the Solver’s initial solution, whereas Verifier2 evaluates the Corrector’s refined solution. Because these two inputs exhibit marked differences in surface form and style,we explicitly split the Verifier into Verifier1 and Verifier2 to ensure that the Verifier is adequately exposed to both types of data. Analogously, the Corrector is partitioned into Corrector1 and Corrector2.

All roles during reasoning for the V-C reasoning system are instantiated within the first five steps. From $t_6$ onward the Verifier–Corrector loop alternates between Verifier2 and Corrector2 until a stopping criterion is met. Consequently, each problem is constrained to a maximum of five rollout iterations during the training phase.
 
\subsection{Agentic Verifiable Rewards}
In reinforcement learning, the final reward is often assigned based on the outcome of the entire trajectory. In multi-agent settings, however, this practice can introduce substantial noise. Consider a three-stage trajectory $\tau = (s, b, c)$, where $s$ is the solution generated by the Solver, $b$ is the bug report produced by the Verifier, and $c$ is the refined result provided by the Corrector. Suppose that the solution $s$ is correct, but the Verifier erroneously identifies errors in it, and the refined result $c$ is also correct. In such a scenario, the reward for the entire trajectory is determined by the reward associated with $c$. Consequently, $b$ is assigned a positive reward due to the correctness of $c$. This outcome is evidently problematic, as the Verifier's incorrect judgment should not warrant a positive reward. To address this issue, we define agent-specific, verifiable rewards computed against a reference answer, rather than relying on downstream outcomes.

\textbf{Reward for the Solver}\\
The reward for the Solver is determined based on the accuracy of the solution $s$ in comparison to a reference answer. Specifically, if $s$ matches the reference answer, the Solver receives a  positive reward; otherwise, a negative reward.

\textbf{Reward for Corrector1/Corrector2}\\
The reward for Corrector1 and Corrector2 is calculated in a manner analogous to that of the Solver. If the refined solution matches the reference answer, the Corrector receives a positive reward; otherwise, a negative reward.

\textbf{Reward for Verifier1/Verifier2}\\
The reward for Verifier1 and Verifier2 is contingent on their ability to correctly identify issues in the solution. If a Verifier incorrectly flags a correct solution as erroneous, the reward is negative; otherwise, positive. Conversely, if a Verifier correctly identifies issues in an incorrect solution, the reward is positive; otherwise negative. The correctness of the solution is determined by its associated reward.

\subsection{Training with Agentic Pipeline Parallelism}
\begin{figure}[ht]
  \centering
  \includegraphics[width=0.95\linewidth]{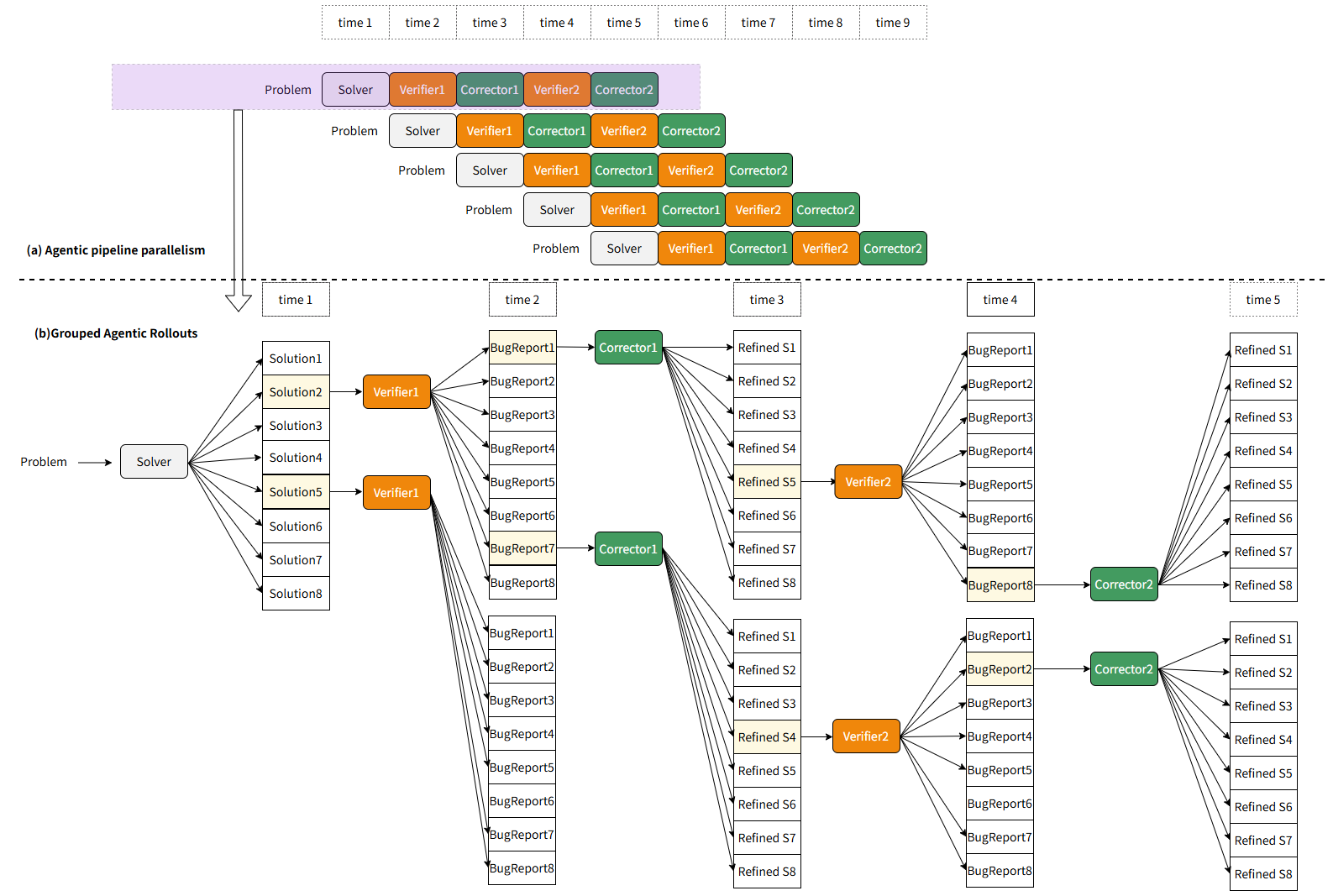}
  \caption{Illustration of agentic pipeline parallelism and grouped rollouts.}
  \label{fig:agentic_rollout}
\end{figure}
Assuming that the maximum output length of the LLM is denoted as \( L \), and that a single sample is processed sequentially by up to \( n \) agents, the total output length for a single trajectory can reach up to \( n \times L \). In this study, we set \( n = 5 \) and \( L = 64\text{k} \), resulting in a maximum potential output length of 320k tokens. However, generating such extended trajectories introduces inefficiencies, particularly when addressing the long-tail distribution of sequence lengths.

Inspired by pipeline parallelism~\citep{huang2019gpipe}, we propose a novel approach that decomposes the training process into a pipeline at the granularity of individual agents. In this framework, once an agent completes its decoding process, its output is immediately added to the training queue, eliminating the need to wait for the completion of the entire trajectory. As depicted in Figure \ref{fig:agentic_rollout}(a), at each time step \(t\), the inputs to the Rollout module are from two sources: the original problem set and the output generated by the preceding agent. The results produced by all agents at each time step are aggregated and utilized for training. This approach significantly reduces the latency between trajectory generation and training. Furthermore, we incorporate segment rollout~\citep{du2025ulorl} within the single-agent decoding process to further enhance efficiency.

\textbf{Grouped Agentic Rollouts}
In the GRPO framework, samples within the same group are decoded based on a shared input, ensuring mutual comparability. This design allows the model to more effectively discern subtle differences between high-quality and suboptimal responses. Consequently, it is crucial to ensure that the samples corresponding to each agent are grouped according to the same input. To address this requirement, we introduce Grouped Agentic Rollouts.

As depicted in Figure \ref{fig:agentic_rollout}(b), for a given problem, the initial agent, referred to as the Solver, generates a set of solutions (the group size is 8). Each subsequent agent selects two outputs from the preceding agent as inputs and produces a group of outputs for each selected input. It is important to note that the Corrector exclusively samples outputs that contain identified errors from the Verifier. If no such outputs are available, the process terminates early.

\subsection{Agentic Sampling Strategies}
\begin{figure}[ht]
  \centering
  \includegraphics[width=0.6\linewidth]{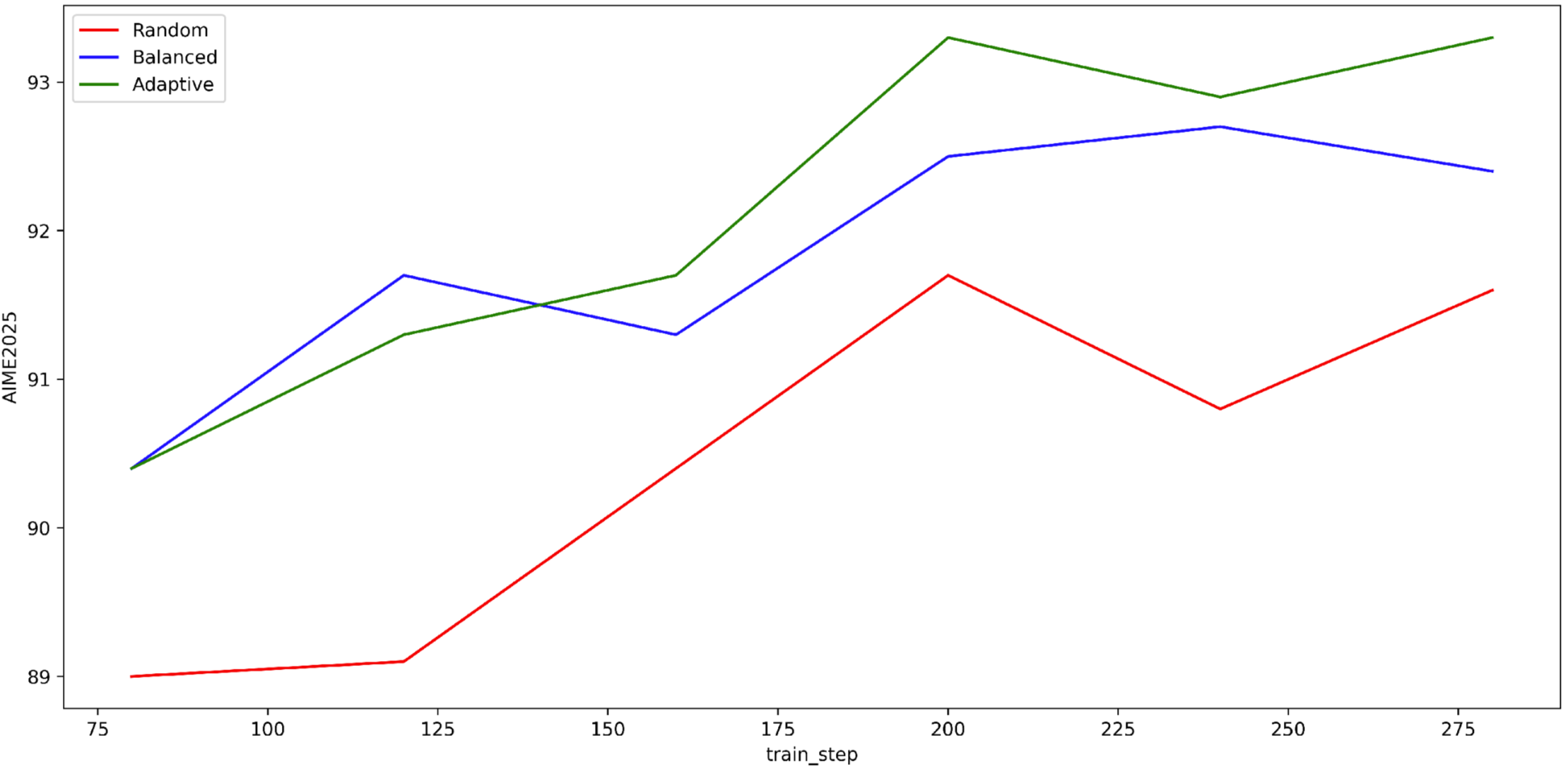}
  \caption{Evaluation results on the AIME-2025 benchmark for different sampling strategies.}
  \label{fig:sampling_strategy_results}
\end{figure}

As mentioned above, each agent samples two instances from its preceding agent. We explore three sampling strategies.

\textbf{Random Sampling (Random)} Randomly sampling two outputs from preceding agent.

\textbf{Negative-Positive Balanced Sampling (Balanced)} Sampling an equal number of positive and negative outputs from preceding agent, where samples with a reward of 1 are considered positive, and those with a reward of 0 are considered negative.

\textbf{Negative-Positive Adaptive Sampling (Adaptive)} The Verifier prioritizes sampling outputs with a reward of 0 (i.e., incorrect solution) from Corrector or Solver, while the Corrector prioritizes sampling outputs with a reward of 1 (i.e., correctly identified errors) from the Verifier.

We conducted experiments based on Qwen3-30B-A3B-Thinking-2507, where all settings were kept consistent except for sampling strategies. We use the V-C reasoning process illustrated in Figure \ref{fig:v-c agents} to evaluate the model's performance on AIME2025. Figure \ref{fig:sampling_strategy_results} presents the results, indicate that the adaptive strategy achieves the best performance, followed by the balanced strategy.

\begin{figure}[ht]
  \centering
  \includegraphics[width=0.9\linewidth]{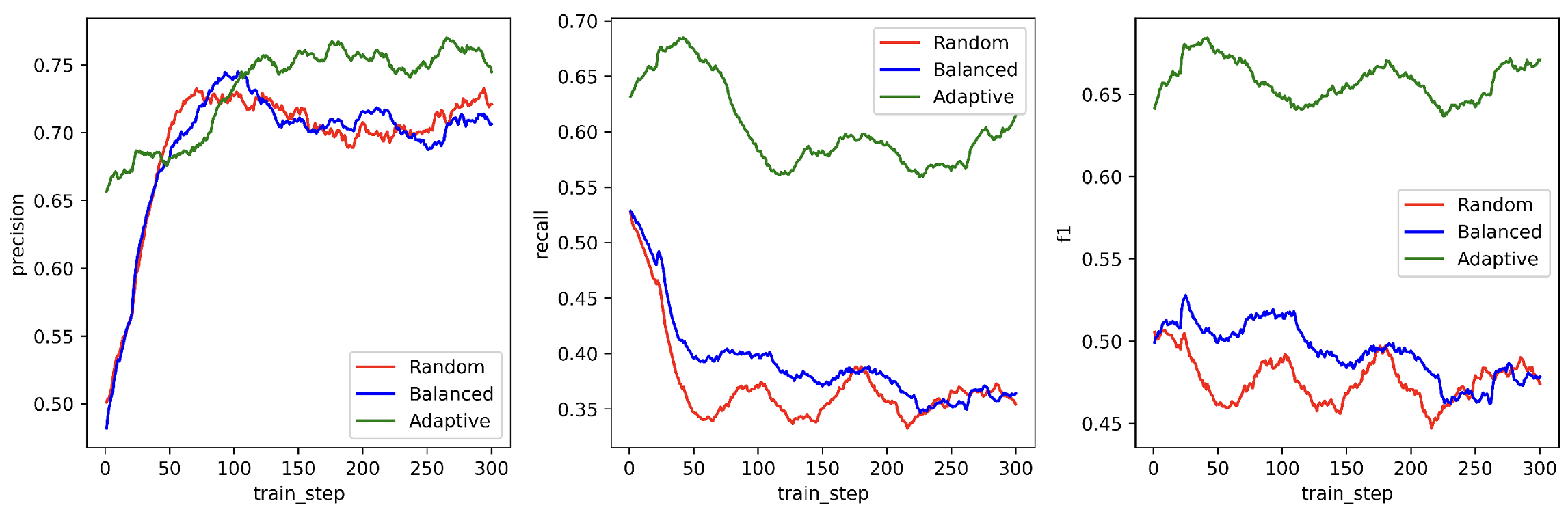}
  \caption{The training dynamics of the Verifier's performance for error detection.}
  \label{fig:dynamics_of_errors}
\end{figure}

As shown in Figure \ref{fig:rl_pipline}, the Corrector within the V-C reasoning system is only capable of addressing errors when the Verifier identifies mistakes in the solutions. Therefore, the Verifier's ability to accurately detect erroneous samples is critical to this system. To evaluate this capability, we track the training dynamics of the Verifier's error detection performance throughout the training process (see Figure \ref{fig:dynamics_of_errors}). It can be observed that the Adaptive method significantly outperforms the other two methods in both accuracy and recall, which explains why this approach achieves the best performance. Based on these findings, the adaptive sampling method is employed as the default approach in subsequent experiments.

\section{Experiments}
\subsection{Training Details}
In this section, we conducted experiments on Qwen3-30B-A3B-Thinking-2507 to verify the effectiveness of MarsRL. Experimental settings are represented as follows.

\textbf{Hyperparameter Settings}
All the agents in our approach share the same hyperparameters. For optimization, we utilize the AdamW optimizer~\citep{zhang2018international} with a constant learning rate of \(1 \times 10^{-6}\). During rollout, the prompt batch size is set to 128, and we sample 8 responses for each prompt. The sampling temperature is set to 0.85, with \(\text{top\_p}=1.0\) and \(\text{top\_k}=-1\). The maximum response length is set to 64k tokens, divided into a maximum of 4 segments, with each segment containing 16k tokens. The maximum input length is set to 32k.

\textbf{Evaluation Setup}
For evaluation, we use the AIME-2025 and BeyondAIME~\citep{yu2025dapo} datasets as benchmarks. Each evaluation set is repeated 32 times, and we report the average score (\(\text{avg@32}\)) to ensure result stability. Fore each agent in the reasoning system, the inference hyperparameters are set to a sampling temperature of 0.6, topp of 0.95 and topk of -1. The maximum response length is set to 64k tokens, and the maximum input length is set to 32k. The evaluation scripts are publicly available\footnote{\url{https://github.com/liushulinle/MarsRL}}.

\subsection{Overall Results}
\begin{table}[!t]
    \centering
        \begin{tabular}{l|c|c|c|c}
            \hline
            \multirow{2}*{\textbf{Model}}  & \multicolumn{2}{c|}{\textbf{AIME-2025}}& \multicolumn{2}{|c}{\textbf{BeyondAIME}}\\
            \cline{2-5}
            & Solver & Reasoning System & Solver & Reasoning System\\
            \midrule
            Qwen3-A3B & 73.5 & 69.7 & 50.7 & 47.6\\
            Qwen3-A3B-Thinking-2507 & 86.5 & 85.6 & 64.9&63.3 \\
            Qwen3-A22B-Thinking-2507 & 92.3 & 91.2 & 70.6& 70.3\\
            Deepsek V3.1-Think & 86.2 & 88.3 & 71.3& 72.0\\
            \hline
            \hline 
            MarsRL-A3B-Thinking-2507 & 91.1 & 93.3 & 70.2& 73.8\\
            \hline
        \end{tabular}
    \caption{Overall results for MarsRL trained on Qwen3-A3B-Thinking-2507. All metrics are from our evaluation.}
    \label{tab:overall}
\end{table}

Table \ref{tab:overall} shows the overall results. Models in the first group are open-sourced models, and the model in the second group is tuned via MarsRL based on Qwen3-30B-A3B-Thinking-2507. From the table, we can make the following observations.

(1) For most models in the first group, the V-C reasoning system does not outperform the Solver used in isolation. An analysis of the data indicates that, although these models follow instructions effectively, they exhibit limited capacity for detecting and correcting errors. 

(2) Following tuning with MarsRL, both the Solver and the reasoning system exhibited substantial performance improvements. On the AIME2025 dataset, the Solver improved from 86.5\% to 91.1\%, while the reasoning system increased from 85.6\% to 93.3\%. On the BeyondAIME dataset, the Solver rose from 64.9\% to 70.2\%, and reasoning system advanced from 63.3\% to 73.8\%. These results provide strong evidence for the effectiveness of MarsRL.

\subsection{Analysis of the Factors Contributing to the Solver's Improvements}

\begin{table}[!t]
    \centering
        \begin{tabular}{l|c|c|c|c}
            \hline
            \multirow{2}*{\textbf{Approach}}  & \multicolumn{2}{c|}{\textbf{AIME-2025}}& \multicolumn{2}{|c}{\textbf{BeyondAIME}}\\
            \cline{2-5}
            & Solver & Reasoning System & Solver & Reasoning System\\
            \midrule
            Qwen3-A3B-Thinking-2507 & 86.5 & 85.6 & 64.9&63.3 \\
            \hline
            \hline
            MarsRL-S & 89.5 & 90.8 & 67.3 & 66.0 \\
            MarsRL-VC & 90.4 & 91.7 & 69.0 & 71.1 \\
            \hline
        \end{tabular}
    \caption{Results of models trained using MarsRL-S and MarsRL-VC.}
    \label{tab:invest_solver}
\end{table}

As mentioned above, the Solver also achieved significant improvements. In this subsection, we investigate whether the observed improvement is attributed to the training of the Solver itself or the training of the Verifier and Corrector. To achieve this goal, we implemented two variant versions based on MarsRL, where only specific agents were updated during training:
\begin{itemize}
    \item \textbf{MarsRL-S} The model is trained exclusively on samples generated by the Solver. Under this configuration, MarsRL-S reduces to UloRL~\citep{du2025ulorl}, a framework for single-model training with long outputs.
    \item \textbf{MarsRL-VC} The model is trained exclusively on samples generated by the Verifier and Corrector; samples from the Solver are excluded during training.
\end{itemize}

\begin{figure}[ht]
  \centering
  \includegraphics[width=0.9\linewidth]{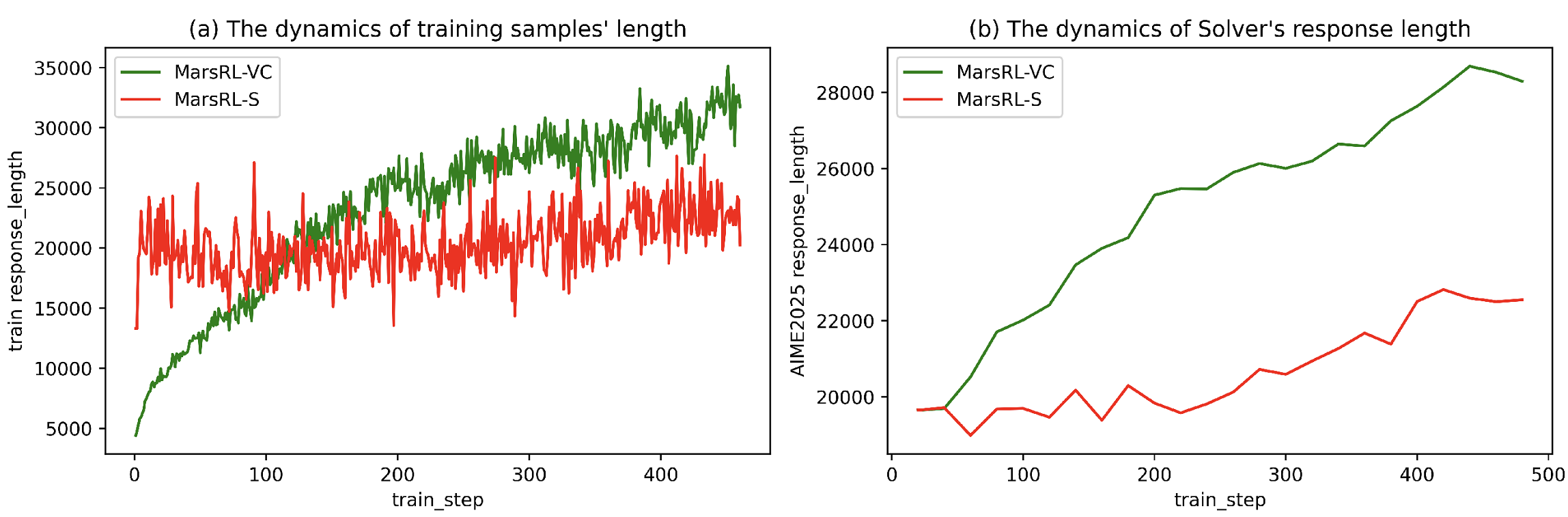}
  \caption{The dynamics of response lengths in models trained using MarsRL-S and MarsRL-VC.}
  \label{fig:solver_invest}
\end{figure}

Table \ref{tab:invest_solver} reports the results. Interestingly, the Solver in MarsRL-VC outperforms its counterpart in MarsRL-S. This result suggests that training the Verifier and Corrector can be more effective for enhancing Solver performance than directly training the Solver. A plausible explanation for this finding is that the model initially lacks adequate verification and correction capabilities. Through training with MarsRL-VC, these capabilities are substantially improved, and this enhancement appears to generalize to the Solver. This hypothesis is further supported by an analysis of changes in the Solver's output length, which is assumed to serve as a proxy for the model's depth of reasoning.

Figure \ref{fig:solver_invest}(a) illustrates the dynamics of sample length involved in model training, and Figure \ref{fig:solver_invest}(b) shows the dynamics of the Solver's average response length. From the figure, we observe that the initial response length of the Verifier and Corrector is only 5k\footnote{Since MarsRL-VC only trains the Verifier and Corrector, its training length refers to the output length of the Verifier and Corrector.}, which is significantly shorter than the Solver's initial length of 19k. This indicates that the model lacks sufficient depth of reasoning for these two tasks. After training, the response length of MarsRL-VC increase rapidly, growing from 5k to 30k, thereby greatly enhancing the model's depth of reasoning. This increase in reasoning depth also generalizes to the Solver's capabilities. Despite not being directly trained, the Solver's output length grows rapidly from 19k to 28k. In contrast, when only the Solver is trained, the model's output length increases very slowly, rising only from 19k to approximately 23k, indicating that the depth of reasoning does not improve significantly. These phenomena provide an explanation for the experimental results presented in Table \ref{tab:invest_solver}. 

\subsection{Evaluating the Generalization Capabilities of Trained V-C Agents}
To assess the generalizability of the Verifier and Corrector, we pair our trained Verifier and Corrector with open-source Solvers. As shown in Table \ref{tab:generalization}, when we replace the Solver with open-source models, the resulting reasoning system consistently achieves substantial improvements over each Solver used in isolation. These findings indicate that the Verifier and Corrector trained with MarsRL generalize effectively across different Solvers.

\begin{table}[!t]
    \centering
        \begin{tabular}{l|c|c|c|c}
            \hline
            \multirow{2}*{\textbf{Solver}}  & \multicolumn{2}{c|}{\textbf{AIME-2025}}& \multicolumn{2}{|c}{\textbf{BeyondAIME}}\\
            \cline{2-5}
            & Solver & Reasoning System & Solver & Reasoning System\\
            \midrule
            Qwen3-A3B-Thinking-2507 & 86.5 & 91.7 & 64.9&71.6 \\
            Qwen3-A22B-Thinking-2507 & 92.3 & 93.3 & 70.6 & 73.3 \\
            DeepSeek V3.1-Think & 86.2 & 91.2 & 71.3 & 74.1 \\
            \hline
        \end{tabular}
    \caption{Results of replacing the Solver by open-source models.}
    \label{tab:generalization}
\end{table}

\section{Conclusions}
In this work, we have introduced MarsRL, a novel reinforcement learning framework designed to advance multi-agent reasoning systems by addressing the challenges of reward noise and training efficiency. By assigning individualized rewards to each agent, MarsRL effectively decouples credit assignment across the Solver, Verifier, and Corrector roles, ensuring more precise and role-specific optimization. Additionally, the adoption of pipeline parallelism significantly enhances training efficiency, enabling the processing of long trajectories without incurring prohibitive computational costs. Our experimental results demonstrate the efficacy of MarsRL in improving reasoning performance across complex tasks. Specifically, applying MarsRL to the Qwen3-30B-A3B-Thinking-2507 model yielded substantial performance gains on AIME2025 and BeyondAIME benchmarks, surpassing even larger models in accuracy. These findings underscore the potential of MarsRL to optimize multi-agent reasoning systems.  
\clearpage
\bibliography{colm2024_conference}
\bibliographystyle{colm2024_conference}

\clearpage
\end{document}